\documentclass{article}

\usepackage{microtype}
\usepackage{graphicx}
\usepackage{subcaption}
\usepackage{booktabs} 
\usepackage{tabularx}
\usepackage{ragged2e} 
\usepackage[table]{xcolor}
\usepackage{multirow}
\definecolor{groupgray}{gray}{0.95}
\usepackage{longtable}
\usepackage{array}

\usepackage[breaklinks=true]{hyperref}

\renewcommand{\em}{\it}

\newcommand{\ignore}[1]{}

\def\cfigure[#1,#2,#3]{
\begin{figure}
\vspace*{0mm}
\begin{center}

\includegraphics[width=3in]{#1} 
 
\vspace*{-3mm}\caption[]{#2
} \label{#3}
 
\vspace*{-5mm}
\end{center}
\end{figure}}

\def\cfigurefour[#1,#2,#3]{
\begin{figure}
\vspace*{0mm}
\begin{center}

\includegraphics[width=4in]{#1} 
 
\vspace*{-3mm}\caption[]{#2
} \label{#3}
 
\vspace*{-5mm}
\end{center}
\end{figure}}

\def\cfiguretemp[#1,#2,#3]{
\begin{figure}
\vspace*{0mm}
\begin{center}

\includegraphics[width=3.5in]{#1} 
 
\vspace*{-3mm}\caption[]{#2
} \label{#3}
 
\vspace*{-5mm}
\end{center}
\vspace*{-2mm}
\end{figure}}

\def\wfigure[#1,#2,#3]{
\begin{figure*}
\vspace*{0mm}
\begin{center}
 \includegraphics[width=\textwidth]{#1} 
 \vspace*{-3mm}\caption[]{#2
} \label{#3}
 
\end{center}
\end{figure*}}

\def\threefigure[#1,#2,#3,#4,#5]{
\begin{figure*}
\vspace*{0mm}
\begin{center}

\begin{tabular}{ccc}
\includegraphics[width=2in]{#1} & \includegraphics[width=2in]{#2} &  \includegraphics[width=2in]{#3} \\
(a) & (b) & (c) \\
\end{tabular}

\vspace*{-3mm}\caption[]{#4
} \label{#5}

\vspace*{-5mm}
\end{center}
\vspace*{-2mm}
\end{figure*}}

\def\dcfigure[#1,#2,#3,#4,#5,#6]{
{
\begin{figure*}
\begin{center}
\begin{minipage}[c]{\columnwidth}{
\includegraphics[width=\columnwidth]{#1} 
\vspace*{0mm}\caption[]{#2} \label{#3} \
}\end{minipage}\hspace*{\columnsep}\
\begin{minipage}[c]{\columnwidth}{
\includegraphics[width=\columnwidth]{#4} 
\vspace*{0mm}\caption[]{#5}\label{#6} \
}\end{minipage}
\end{center}
\end{figure*}
}
}

\def\tableByTable[#1,#2,#3,#4,#5,#6]{
{
\begin{table*}
\begin{center}
\begin{minipage}[c]{3in}{
\centering
{#1}
\vspace*{0mm}\tabcaption[]{#2}\label{#3} \
}\end{minipage}\hspace*{\columnsep}\
\begin{minipage}[c]{3in}{
\centering
{#4}
\vspace*{0mm}\tabcaption[]{#5}\label{#6} \
}\end{minipage}
\end{center}
\end{table*}
}
}

\def\figureByTable[#1,#2,#3,#4,#5,#6]{
{
\begin{figure*}
\begin{center}
\begin{minipage}[c]{3in}{
\centering
\includegraphics[width=\textwidth]{#1}
\vspace*{0mm}\figcaption[]{#2} \label{#3} \
}\end{minipage}\hspace*{\columnsep}\
\begin{minipage}[c]{3.3in}{
\centering
{#4}
\vspace*{0mm}\tabcaption[]{#5}\label{#6} \
}\end{minipage}
\end{center}
\end{figure*}
}
}

\def\tableByFigure[#1,#2,#3,#4,#5,#6]{
{
\begin{figure*}
\begin{center}
\begin{minipage}[c]{4.3in}{
\centering
{#1}
\vspace*{0mm}\tabcaption[]{#2} \label{#3} \
}\end{minipage}\hspace*{\columnsep}\
\begin{minipage}[c]{2.2in}{
\centering
\includegraphics[width=\textwidth]{#4}
\vspace*{-0.35in}\caption[]{#5}\label{#6} \
}\end{minipage}
\end{center}
\end{figure*}
}
}

\def\doublecfigure[#1,#2,#3,#4]{
{
\begin{figure}
\begin{center}
\begin{minipage}[c]{1.5in}{
\begin{center}
\includegraphics[width=1.5in]{#1}
\end{center}
}\end{minipage}\hspace*{1em}\
\begin{minipage}[c]{1.5in}{
\begin{center}
\includegraphics[width=1.5in]{#2}
\end{center}
}\end{minipage}
\vspace*{0mm}\caption[]{#3} \label{#4} \
\end{center}
\end{figure}
}
}

\def\qcfigure[#1,#2,#3,#4,#5,#6]{
{
\begin{figure*}
\vspace*{0.2in}\
\begin{center}
\begin{minipage}[c]{3in}{
\includegraphics[width=3in]{#1} 
\vspace*{-3mm}
}
\end{minipage}\hspace*{0.5in}\
\begin{minipage}[c]{3in}{
\includegraphics[width=3in]{#2} 
\vspace*{-3mm}
}\end{minipage}

\begin{minipage}[c]{3in}{
\includegraphics[width=3in]{#3} 
\vspace*{-3mm}
}
\end{minipage}\hspace*{0.5in}\
\begin{minipage}[c]{3in}{
\includegraphics[width=3in]{#4} 
\vspace*{-3mm}
}\end{minipage}
\end{center}
\caption[]{#5}\label{#6}
\end{figure*}
}
}

\def\twfigure[#1,#2,#3,#4,#5]{
{
\begin{figure*}
\vspace*{0.2in}\
\begin{center}
\begin{minipage}[c]{6.5in}{
\includegraphics[width=6.5in]{#1} 
\vspace*{-3mm}
}
\end{minipage}

\begin{minipage}[c]{6.5in}{
\includegraphics[width=6.5in]{#2} 
\vspace*{-3mm}
}\end{minipage}

\begin{minipage}[c]{6.5in}{
\includegraphics[width=6.5in]{#3} 
\vspace*{-3mm}
}
\end{minipage}
\end{center}
\caption[]{#4}\label{#5}
\end{figure*}
}
}

\def\dwfigure[#1,#2,#3,#4]{
{
\begin{figure*}
\vspace*{0.2in}\
\begin{center}
\begin{minipage}[c]{6.5in}{
\includegraphics[width=6.5in]{#1} 
\vspace*{-3mm}
}
\end{minipage}

\begin{minipage}[c]{6.5in}{
\includegraphics[width=6.5in]{#2} 
\vspace*{-3mm}
}\end{minipage}

\end{center}
\caption[]{#3}\label{#4}
\end{figure*}
}
}

\def\dssfigure[#1,#2,#3,#4,#5,#6]{
{
\begin{figure*}
\vspace*{0.2in}\
\begin{center}
\begin{minipage}[c]{4in}{
\includegraphics[width=4in]{#1}
\vspace*{-3mm}\caption[]{#2} \label{#3} \
}\end{minipage}\hspace*{0.5in}\
\begin{minipage}[c]{2in}{
\includegraphics[width=2in]{#4}
\vspace*{-3mm}\caption[]{#5}\label{#6} \
}\end{minipage}
\end{center}
\vspace*{-0.4in}\
\end{figure*}
}
}

\def\dsfigure[#1,#2,#3,#4,#5,#6]{
{
\begin{figure*}
\vspace*{0.2in}\
\begin{center}
\begin{minipage}[c]{3in}{
\includegraphics[width=3in]{#1}
\vspace*{-3mm}\caption[]{#2} \label{#3} \
}\end{minipage}\hspace*{0.5in}\
\begin{minipage}[c]{3in}{
\hspace*{0.5in}\
\includegraphics[height=3in]{#4}
\vspace*{-3mm}\caption[]{#5}\label{#6} \
}\end{minipage}
\end{center}
\vspace*{-0.4in}\
\end{figure*}
}
}

\def\dsyfigure[#1,#2,#3,#4,#5,#6]{
{
\begin{figure*}
\vspace*{0.2in}\
\begin{center}
\begin{minipage}[c]{2.5in}{
\includegraphics[height=2.5in]{#1}
\vspace*{-3mm}\caption[]{#2} \label{#3} \
}\end{minipage}\hspace*{0.5in}\
\begin{minipage}[c]{2.5in}{
\includegraphics[height=2.5in]{#4}
\vspace*{-3mm}\caption[]{#5}\label{#6} \
}\end{minipage}
\end{center}
\vspace*{-0.4in}\
\end{figure*}
}
}

\def\dyfigure[#1,#2,#3,#4,#5,#6]{
{
\begin{figure*}
\vspace*{0.2in}\
\begin{center}
\begin{minipage}[c]{3in}{
\includegraphics[height=3in]{#1} 
\vspace*{-3mm}\caption[]{#2} \label{#3} \
}\end{minipage}\hspace*{0.5in}\
\begin{minipage}[c]{3in}{
\includegraphics[height=3in]{#4} 
\vspace*{-3mm}\caption[]{#5}\label{#6} \
}\end{minipage}
\end{center}
\vspace*{-0.4in}\
\end{figure*}
}
}

\def\dyoldfigure[#1,#2,#3,#4,#5,#6]{
{
\begin{figure*}
\vspace*{0.2in}\
\begin{center}
\begin{minipage}[c]{3in}{
\epsfysize=2.0in\
\hspace{0.5in}\
\epsfbox{#1}
\vspace*{-3mm}\caption[]{#2} \label{#3} \
}\end{minipage}\hspace*{0.25in}\
\begin{minipage}[c]{3in}{
\epsfysize=2.0in\
\hspace{0.5in}\
\epsfbox{#4}
\vspace*{-3mm}\caption[]{#5}\label{#6} \
}\end{minipage}
\end{center}
\vspace*{-0.4in}\
\end{figure*}
}
}

\def\cfiguredouble[#1,#2,#3,#4]{
\begin{figure}
\vspace*{0.2in}\
\begin{center}
\begin{minipage}[c]{1.5in}{
\epsfxsize=1.5in\
\epsfbox{#1}
}\end{minipage}\hspace*{0.1in}\
\begin{minipage}[c]{1.5in}{
\epsfxsize=1.5in\
\vspace{0.1in}\epsfbox{#2}
}\end{minipage}\vspace*{-0.10in} \caption[]{#3}\label{#4}
\end{center}
\vspace*{-0.4in}\
\end{figure}
}

\def\wpfigure[#1,#2,#3,#4]{
\begin{figure*}
\vspace*{4mm}
\begin{center}

\includegraphics[width=#4]{#1} 

\vspace*{-3mm}\caption[]{#2
} \label{#3}

\vspace*{-5mm}
\end{center}
\end{figure*}}

\def\wprfigure[#1,#2,#3,#4,#5]{
\begin{figure*}
\vspace*{4mm}
\begin{center}

\includegraphics[width=#4, angle=#5]{#1} 

\vspace*{-3mm}\caption[]{#2
} \label{#3}

\vspace*{-5mm}
\end{center}
\end{figure*}}

\def\DoubleFigureWSlide[#1,#2,#3,#4,#5,#6,#7,#8,#9]{
\begin{figure*}
\vspace*{#9}
\begin{center}
\begin{minipage}{#4}
\includegraphics[width=#4]{#1}
\vspace*{-3mm}\caption{#2
}\label{#3}
\end{minipage}
\hspace{2em}
\begin{minipage}{#8}
\includegraphics[width=#8]{#5}
\vspace*{-3mm}\caption{#6
}\label{#7}
\end{minipage}
\vspace*{-5mm}
\end{center}
\end{figure*}
}

\def\DoubleFigureW[#1,#2,#3,#4,#5,#6,#7,#8]{
\begin{figure*}
\vspace*{0in}
\begin{center}
\begin{minipage}{#4}
\includegraphics[width=#4]{#1}
\vspace*{-3mm}\caption{#2
}\label{#3}
\end{minipage}
\hspace{2em}
\begin{minipage}{#8}
\includegraphics[width=#8]{#5}
\vspace*{-3mm}\caption{#6
}\label{#7}
\end{minipage}
\vspace*{-5mm}
\end{center}
\end{figure*}
}

\def\DoubleFigureWHack[#1,#2,#3,#4,#5,#6,#7,#8]{
\begin{figure*}
\vspace*{0in}
\begin{center}
\begin{minipage}{3in}
\includegraphics[width=#4]{#1}
\vspace*{-3mm}\caption{#2
}\label{#3}
\end{minipage}
\hspace{2em}
\begin{minipage}{3in}
\includegraphics[width=#8]{#5}
\vspace*{-3mm}\caption{#6
}\label{#7}
\end{minipage}
\vspace*{-5mm}
\end{center}
\end{figure*}
}

\def\ddcfigure[#1,#2,#3,#4]{
\begin{figure*}
\vspace*{0.2in}\
\begin{center}
\begin{minipage}[c]{\columnwidth}{
\includegraphics[width=\columnwidth]{#1} 
}\end{minipage}\hspace{0.5in}\
\begin{minipage}[c]{\columnwidth}{
\includegraphics[width=\columnwidth]{#2} 
}\end{minipage} \caption[]{#3}\label{#4}
\end{center}
\end{figure*}
}

\def\ddcfigureSlide[#1,#2,#3,#4,#5]{
\begin{figure*}
\vspace*{#5}\
\begin{center}
\begin{minipage}[c]{3in}{
\includegraphics[height=3in]{#1} 
}\end{minipage}\hspace{0.5in}\
\begin{minipage}[c]{3in}{
\includegraphics[height=3in]{#2} 
}\end{minipage}\vspace*{-0.10in} \caption[]{#3}\label{#4}
\end{center}
\vspace*{-0.4in}\
\end{figure*}
}

\def\cxfigure[#1,#2,#3]{
\begin{figure}
\vspace*{4mm}
\begin{center}
 
\epsfxsize=2.5in\
\epsfbox{#1}\
 
\vspace*{-0.10in}\caption[]{#2
} \label{#3}
 
\vspace*{-5mm}
\end{center}
\vspace*{-2mm}
\end{figure}}

\newcommand{\beforecaption}{\vspace{-.15cm}\begin{spacing}{0.85}}
\newcommand{\aftercaption}{\vspace{-.45cm}\end{spacing}}

\newcommand{\eg}{\textit{e.g.}}
\newcommand{\ie}{\textit{i.e.}}

\newcommand{\bench}{SourceBench}
\newcommand{\semantic}{\texttt{CR}}
\newcommand{\fact}{\texttt{FA}}
\newcommand{\obj}{\texttt{NE}}
\newcommand{\fresh}{\texttt{FR}}
\newcommand{\org}{\texttt{OA}}
\newcommand{\auth}{\texttt{AA}}
\newcommand{\domain}{\texttt{DA}}
\newcommand{\ad}{\texttt{LC}}

\newcommand{\boldunderparagraph}[1]{\vspace*{0.5ex}\noindent\underline{\textbf{#1}}}

\usepackage{tikz}

\newcommand{\myitem}[1]{\item \textbf{#1}}

\usepackage[preprint]{icml2026}
\usepackage{xurl}

\usepackage{amsmath}
\usepackage{amssymb}
\usepackage{mathtools}
\usepackage{amsthm}

\usepackage[capitalize,noabbrev]{cleveref}

\theoremstyle{plain}

\theoremstyle{definition}

\theoremstyle{remark}

\usepackage[textsize=tiny]{todonotes}

\icmltitlerunning{\bench: Can AI Answers Reference Quality Web Sources?}

\begin{document}

\twocolumn[
  \icmltitle{\bench: Can AI Answers Reference Quality Web Sources?}



  \icmlsetsymbol{equal}{*}

  \begin{icmlauthorlist}
    \icmlauthor{Hexi Jin}{equal,ucsd}
    \icmlauthor{Stephen Liu}{equal,ucsd}
    \icmlauthor{Yuheng Li}{equal,ucsd}
    \icmlauthor{Simran Malik}{equal,ucsd}
    \icmlauthor{Yiying Zhang}{ucsd,comp}
  \end{icmlauthorlist}

  \icmlaffiliation{ucsd}{University of California, San Diego}
  \icmlaffiliation{comp}{GenseeAI Inc.}

  \icmlcorrespondingauthor{Yiying Zhang}{yiying@ucsd.edu}

  \icmlkeywords{Machine Learning, ICML}

  \vskip 0.3in
]



\printAffiliationsAndNotice{}  

\begin{abstract}

Large language models (LLMs) increasingly answer queries by citing web sources, but existing evaluations emphasize answer correctness rather than evidence quality. We introduce \bench, a benchmark for measuring the quality of cited web sources across 100 real-world queries spanning informational, factual, argumentative, social, and shopping intents. \bench\ uses an eight-metric framework covering content quality (content relevance, factual accuracy, objectivity) and page-level signals (\eg, freshness, authority/accountability, clarity), and includes a human-labeled dataset with a calibrated LLM-based evaluator that matches expert judgments closely. We evaluate eight LLMs, Google Search, and three AI search tools over 3996 cited sources using \bench\ and conduct further experiments to understand the evaluation results. Overall, our work reveals four key new insights that can guide future research in the direction of GenAI and web search.

\end{abstract}

\section{Introduction}

Large Language Models (LLMs) excel at language generation but rely on static parametric memory and can hallucinate~\cite{hoh_benchmark, Li2024}. Web search mitigates these limitations by grounding and enriching responses with external, up-to-date evidence. As such, it has become the de facto in many production LLM systems such as ChatGPT~\cite{OpenAI2024SearchGPT}, Gemini~\cite{Gemini2023}, Perplexity~\cite{Perplexity2022}, and Microsoft Copilot~\cite{Microsoft2023Bing}. 

The integration of search in LLMs introduces a new dependency: the quality of sources used by LLMs directly determines the groundedness of the generated response. In high-stakes domains—such as financial analysis, medical inquiries, or legal research—users increasingly rely on cited references for verification. Recent studies indicate that the mere presence of citations significantly increases user trust, even if the underlying sources are not rigorously checked by the user~\cite{Ding2025}. This creates a vulnerability where a search-augmented LLM is subject to the ``Garbage In, Garbage Out'' principle; if an LLM synthesizes information from low-quality, biased, or outdated web pages, the resulting answer remains flawed or misleading.

Despite the importance of web source quality, existing benchmarks and evaluators remain focused on the AI answer text. Benchmarks like HotpotQA~\cite{hotpotqa} and HLE~\cite{HLE} assess the accuracy of the final answer, while retrieval benchmarks like RAGBench~\cite{RAGBench} and evaluators like RAGAS~\cite{RAGAS} focus on relevance ranking rather than source credibility. 
None of these systematically evaluates {\em the web sources themselves} or how users experience them.

In this paper, we present \textbf{\textit{\bench}}, a novel benchmark to evaluate the quality of web sources referenced in AI answers. We argue that the utility of a retrieved source is not defined solely by its textual similarity to the query, but by the holistic experience it offers to the human verifier. As AI systems increasingly function as information curators, users must be able to seamlessly transition from reading a synthesized answer to auditing its provenance. A source that is factually correct but buried under intrusive ads, or one that lacks a clear publication date, imposes a high cognitive load on the user, breaking the trust loop essential for high-stakes information retrieval.

Consequently, we depart from standard relevance-only metrics to propose a comprehensive, multi-facet scoring framework designed from the user's perspective. We evaluate sources across \textit{eight metrics} organized into two categories. 

The first category assesses web sources based on their page {\em content}, including content relevance, factual accuracy, and objectivity. These metrics ensure the information is not only textually aligned with the query but is also substantively correct and free from manipulative bias. For a user, this is the baseline requirement.
%
The second category evaluates ``{\em meta-attributes}'' of web sources, including freshness, organization, author, and domain authority, and layout clarity. We posit that these meta-attributes are the primary drivers of user trust and experience and are as important. 

To systematically evaluate, we introduce a human-aligned automated evaluator. By fine-tuning an LLM on human-labeled samples and utilizing a split-prompt strategy for content and metadata rubrics, we achieve high-fidelity scoring that closely mirrors manual grading.

We utilize \bench\ to conduct a comprehensive study of \textbf{3996 web sources} from \textbf{12 systems} in three categories: search-equipped LLMs (\eg, GPT-5~\cite{openai2025gpt5}, Gemini-3-pro~\cite{google2025gemini3}), traditional SERP (Google), and AI-native search tools (\eg, Exa~\cite{exa_docs}, Tavily~\cite{tavily_docs}). Our findings reveal distinct behavioral profiles: GPT-5 achieves the highest overall scores driven by its superior ability to vet sources for various types of accountability and factuality, while Grok-4.1~\cite{xAI2025Grok41} and Gensee AI search~\cite{Gensee2026} excel in content relevance and achieve the second and third highest scores.

We then conduct a set of experiments to further understand the relationship between AI answer and web search. Through controlled experiments with DeepSeek models~\cite{deepseek_v3_2_2025}, we find that a non-reasoning model with a high-quality search tool outperforms a reasoning model with lower-quality search tool, suggesting that high-quality web source selection can functionally substitute for, and even outperform, advanced internal reasoning steps. At the same time, reasoning is still important, as we find that the highest-performing AI systems rely on sources that significantly diverge from traditional Google Search results, indicating that complex queries require multi-step thinking that go beyond keyword-based search rankings.

This paper makes the following key contributions:

\begin{itemize}
    \myitem{A Multi-Facet Web Source Quality Framework}: Eight distinct rubrics for evaluating the quality of web sources from a user perspective.

\myitem{The \bench\ Dataset and Evaluator}: A diverse dataset of 100 queries and an automated, human-validated evaluation pipeline. 


\myitem{Comprehensive System Analysis}: Evaluation of 12 systems, including eight popular LLMs, three AI-based web search tools, and the Google search.

\myitem{In-depth Analysis of GenAI and Web Search Relationship}: A set of additional experiments that reveal new insights, guiding future researchers and practitioners.

\end{itemize}

We will make \bench\ and its evaluator publicly available upon the acceptance of this paper.

\section{Motivation and Related Work}
\label{sec:motivation}

\subsection{Search and GenAI}
\label{sec:background}

Pre-trained Large Language Models (LLMs) have demonstrated remarkable capabilities in reasoning and natural language generation, yet they suffer from two fundamental limitations inherent to their architecture: static parametric memory and the tendency to hallucinate. An LLM’s knowledge is frozen at its pre-training time, rendering it blind to any events happening afterward or any changes to the training data (\eg, a website corrects itself after the pre-training time). Furthermore, LLMs today still suffer greatly from hallucination and often generate plausible but factually incorrect responses.

Web search acts as the critical enabler to transcend LLMs frozen knowledge base and possible hallucination. By integrating search results into the generation process, LLMs can ground their responses in external evidence. This not only mitigates hallucinations by providing factual context but also allows the model to access up-to-date information without any retraining.
As such, web searches have been integrated in almost all popular LLMs like ChatGPT~\cite{OpenAI2024SearchGPT}, Gemini~\cite{Gemini2023}, Perplexity~\cite{Perplexity2022}, and Bing Copilot~\cite{Microsoft2023Bing}.

Meanwhile, users increasingly rely on cited references in AI answers for verification. In high-stakes domains like financial analysis, medical inquiries, or legal opinions, AI answers are only as valuable as the evidence supporting them. 
A search-augmented LLM follows the {\em ``Garbage In, Garbage Out''} principle: if the LLM retrieves and synthesizes information from low-quality web pages, the resulting answer will be flawed, irrelevant, or outdated. Therefore, the quality of the web source itself is paramount.
Unfortunately, despite the prolificacy of LLM benchmarks, there is no benchmark to directly evaluate web sources in AI answers.




\subsection{Related Benchmarks and Evaluation}
\label{sec:relateed}

Most existing LLM benchmarks focus on the accuracy and quality of LLM answers, such as HotpotQA~\cite{hotpotqa}, HLE~\cite{HLE}, and SWEBench~\cite{SWEBench}.
These benchmarks target the evaluation of LLM generation that does not reference external sources.

Frameworks like RAGAS (Retrieval-Augmented Generation Assessment)~\cite{RAGAS} assess the retrieved information via context precision and context recall. ARES (Automated RAG Evaluation System)~\cite{ARES} refines this by training specialized LLM judges on synthetic datasets to provide more reliable, confidence-aware judgments. Other frameworks like RAGProbe~\cite{RAGProbe} and RAGBench~\cite{RAGBench} focus on specific aspects like automated testing for failure modes and explainability.

While these frameworks effectively measure if an answer is faithful to a given context, none of them evaluates the web sources referenced in the answer. Checking the quality of the citation itself requires a different approach, as we will demonstrate next. 

\section{\bench}
\label{sec:benchmark}

This section introduces \bench, the first benchmark to evaluate the quality of web sources used in AI answers. We first introduce the benchmark requests and then discuss the metrics we design for evaluating source quality. Next, we discuss how we manually label reference sources on these metrics and how we leverage the manual labeling results to design an automated evaluator.

\subsection{Request Collection}
\label{sec:query}




We compiled a diverse dataset of 100 queries by randomly sampling from DebateQA~\cite{debateqa}, HotpotQA~\cite{hotpotqa}, Pinocchios~\cite{pinocchio}, Quora Question Pairs~\cite{quora_kaggle}, and VACOS\_NLQ~\cite{vacos_nlq}. This selection includes e-commerce Q\&A sites like Amazon and eBay, forum posts from Quora and Reddit, and search query logs—which represent real-world user questions as seen in datasets like MS MARCO~\cite{msmarco}. or commercial query logs. This broad selection captures a wide spectrum of user intents, ranging from commercial product comparisons and purchase guidance to informational factual lookups. We summarize these queries' type and sources in Table~\ref{tbl:query}.

\begin{table*}[!t]
\small
  \begin{center}
  \caption{\textbf{Requests Composition in \bench.}}
    \vspace{1em}
  \begin{tabular}{llcl}
    \hline
    \textbf{Type} & \textbf{Source} & \textbf{Count} & \textbf{Primary Intent} \\
    \hline
    Shopping & VA-COS NLQ~\cite{vacos_nlq} & 20 & Product search and purchase guidance \\
    Reasoning & HotpotQA~\cite{hotpotqa} & 20 & Multi-hop factual research and lookups \\
    Factuality & Pinocchios~\cite{pinocchio} & 20 & Fact-checking and claim verification \\
    Argumentative & DebateQA~\cite{debateqa} & 20 & Multi-perspective subjective reasoning \\
    Social Q\&A & Quora~\cite{quora_kaggle} & 20 & Community-based forum inquiries \\
    \hline
    \textbf{Total} & & \textbf{100} & \\
    \hline
  \end{tabular}
  \label{tbl:query}
  \end{center}
\end{table*} 

We anticipate LLMs to answer the queries with both summary text and referenced sources, expressed as URLs. 
Our goal of \bench\ is to evaluate the quality of the referenced sources, not the summary text answer itself. As explained in Section~\ref{sec:relateed}, many benchmarks such as HotpotQA~\cite{hotpotqa} and DebateQA~\cite{debateqa} already evaluate generated text quality, and we do not include this as \bench's metrics. Thus, the input to \bench\ is the user query and URLs 



\subsection{Multi-Facet Source Quality Metrics}
To evaluate the quality of AI sources, we propose a multi-facet scoring framework. To provide a well-rounded assessment of a source's quality, we propose eight metrics (each with numerical score of 1 to 5) organized into two categories, as summarized in Table~\ref{table:rubric}. We design these metrics from the users' perspective, with the goal of giving higher scores to sources that users are most likely to appreciate and read. 

The first dimension is {\em content quality}, which evaluates the content of a source web page. We measure content quality with three metrics: {\em Content Relevance} (\textbf{\semantic}), {\em Factual Accuracy} (\textbf{\fact}), and {\em Objectivity} (\textbf{\obj}). High \semantic, \fact, and \obj \ 
scores mean that the source content directly provides information to answer the user's question, provides information that is verifiable, and objectively delivers the information.

The second dimension is on the {\em meta-feature} of source pages, including {\em Freshness} (\textbf{\fresh}), {\em Ownership} (\textbf{\org}), {\em Author Accountability} (\textbf{\auth}), {\em Domain Authority} (\textbf{\domain}), and {\em Layout Clarity} (\textbf{\ad}). These metrics are evaluated based on metadata of sources instead of their content. They are important to users because for similar content, users trust sources that are more accountable (organization, author, and domain authority), recent (the last time a website is updated or maintained), and clear (clearly laid out and low ad density). 

Specifically, Freshness checks the last time a web page is updated or maintained (\ie, being checked by human for content still being correct). It ensures that AI answers are not based on obsolete data (\eg, deprecated code or outdated regulations).
Accountability and Authority allow users to gauge the credibility of the expertise (\eg, distinguishing a peer-reviewed medical study from an anonymous forum post).
Finally, Layout Clarity evaluates the ``consumability'' of the webpage itself. In the current web ecosystem, users are often directed to ``SEO farms'' saturated with ads and pop-ups. By penalizing such sources, we explicitly value the user's ability to efficiently verify information without fighting through hostile user interface patterns.

\begin{table*}[t]
\caption{{\textbf{\bench\ Rubrics}}. Each metric is scored on a 1--5 scale with \textbf{5} (best) and \textbf{1} (worst) meaning described below.}
\label{table:rubric}
\vskip 0.12in
\begin{center}
\begin{small}
\newcolumntype{Y}{>{\RaggedRight\arraybackslash}X}
\setlength{\tabcolsep}{6pt}

\begin{tabularx}{\textwidth}{@{} >{\RaggedRight\arraybackslash}p{0.05\textwidth}
                        >{\RaggedRight\arraybackslash}p{0.17\textwidth}
                        >{\RaggedRight\arraybackslash}p{0.35\textwidth}
                        Y @{}}
\toprule
\textbf{Metric} & \textbf{Focus} & \textbf{Best Score (5)} & \textbf{Worst Score (1)} \\
\midrule

\textbf{\semantic} &
Content relevance \& problem resolution &
Comprehensive and directly resolves the user need; no further lookup required. &
Irrelevant or mismatched to the query. \\

\addlinespace

\textbf{\fact} &
Factual accuracy \& verifiability &
Claims are verifiable and supported by citations, prioritizing primary sources. &
Contains false or harmful claims; unverifiable assertions that contradict known facts. \\

\addlinespace

\textbf{\obj} &
Neutrality \& objectivity &
Neutral/clinical delivery; avoids emotional manipulation and does not push a single agenda. &
Propagandistic or manipulative tone that distorts judgment. \\

\addlinespace
\hline\addlinespace

\textbf{\fresh} &
Freshness \& maintenance &
Timely and clearly maintained, dated/updated recently or follow query’s time context. &
Legacy/obsolete: time-sensitive info is outdated such that it becomes incorrect or misleading. \\

\addlinespace

\textbf{\auth} &
Author accountability \& expertise &
Named author with verifiable bio/credentials and clear accountability links. &
Deceptive or missing identity signals (fabricated persona; no credible author info). \\

\addlinespace

\textbf{\org} &
Ownership accountability &
Ownership and funding are transparent with clear organizational accountability. &
Ownership is hidden. No contact/organizational info; signals of shell or obfuscation. \\

\addlinespace

\textbf{\domain} &
Domain authority \& institutional reputation &
High-authority domain with established reputation and long-standing trust signals. &
Known scam/disinformation domain or consistently low-credibility reputation. \\

\addlinespace

\textbf{\ad} &
Layout Clarity \& integrity &
Readable with minimal obstruction; ads do not interfere with accessing the main content. &
High ad density, injected ``ghost text''/scraped fragments, or overlays that obstruct reading. \\

\addlinespace
\bottomrule
\end{tabularx}
\end{small}
\end{center}
\vskip -0.10in
\end{table*}




\subsection{Evaluator}

We build our evaluator by first collecting source URLs, then manually label a sampled set of them, and finally develop an LLM-based evaluator with the help of manual label examples.

\subsubsection{Source Collection.}
To establish a human reference of \bench\ scores, 
we collect source URLs from three distinct categories: popular LLMs, traditional Search Engine Results Pages (SERP), and AI search engines. For traditional SERP, we utilize the Google Search API to record the top search result URLs. To reflect modern AI-driven discovery, we also include sources retrieved from AI search engines such as Tavily~\cite{tavily_docs}, Exa~\cite{exa_docs} and Gensee~\cite{Gensee2026}.

when answering the 100 queries in \bench. For SERP, we call the Google search API and record the top-5 search result URLs. For each URL, we scrape the page content. These page contents form the dataset for our evaluator development.


\subsubsection{Human Labeling}
We sample 45 scraped pages and manually grade them on the eight metrics in Table~\ref{table:rubric}. Two graduate students grade each source based on a unified grading scheme and rubrics. Doing so largely reduces human variance and provide credible foundation for building our automated evaluator. During the manual grading period, we also identify ambiguities in our metrics and improve them accordingly.

\begin{table*}[ht]
    \centering
    \small
    \caption{Qwen-Plus Few-shot Alignment with Human Graders ($N=45$). Values represent the Mean Absolute Error (MAE) between Model scores and the average of Human Graders.}
    \label{tab:qwen_human}
    \renewcommand{\arraystretch}{1.2}
    \begin{tabular}{lccccccccc}
    \toprule
    \textbf{Dimension} & \textbf{\semantic} & \textbf{\fact} & \textbf{\obj} & \textbf{\fresh} & \textbf{\org} & \textbf{\auth} & \textbf{\domain} & \textbf{\ad} & \textbf{Avg} \\
    \midrule
    \textbf{MAE} & $0.56$ & $0.37$ & $0.50$ & $0.91$ & $0.54$ & $0.44$ & $0.41$ & $0.46$ & $\mathbf{0.20}$ \\
    \bottomrule
    \end{tabular}
\end{table*}

\subsubsection{LLM Evaluator}


To automate the evaluation process, we employ a few-shot prompting strategy using human-verified labels as exemplars, designing distinct system prompts for each of the eight evaluation rubrics. For input processing, we capture and utilize the full textual content of each scraped web source; our preliminary experiments demonstrated that summarizing or chunking the content significantly degraded scoring accuracy, necessitating the use of the entire document. To accommodate this full-text requirement without exceeding the context window, we partition the rubrics into two content-based and metadata-based, each paired with its own tailored set of few-shot examples. To choose the appropriate few-shot examples for each rubric, we use a fine-tuning process on our curated dataset of 45 human-labeled samples to ensure strict alignment with manual grading standards. Appendix~\ref{appendix:prompts} presents our full prompts. 
Table~\ref{tab:qwen_human} presents the mean error of our LLM-based evaluator and human graders (out of 5). As seen, our automated scoring approach achieves high fidelity, with the model's mean score deviations consistently remaining within 0.5 points of the human consensus across all rubrics.

\section{Evaluation Results}

\subsection{Evaluating Systems}

There are three primary ways of getting answers and finding web sources: 1) using LLM-based answers, 2) using traditional SERP-based search like Google, and 3) using AI-based web search tools like Exa~\cite{exa_docs} and Tavily~\cite{tavily_docs}. They represent different levels of LLM/search: LLM-primary search-secondary, search-primary, and search-primary LLM-secondary.
To understand how well they do in finding quality sources that are valuable to users, we conduct a comprehensive evaluation of 12 systems in the three categories using \bench.
They include GPT-5, GPT-4o, Gemini-3-Pro, Gemini-3-flash, Gemini-2.5-flash, Grok-4.1-Fast, Claude-Sonnet-4.5, Perplexity-Sonar-Pro, Google search, Exa AI Search, Tavily AI Search, and Gensee AI Search~\cite{Gensee2026}.
For the LLMs, we extract the URLs that they output in their final answer. URLs visited (\eg, during intermediate reasoning steps) but not used in the final answer are not included.
For Exa, Tavily, and Gensee, we directly call their developer APIs, which output a list of URLs.
For Google, we call its standard Google web search API, which returns a ranked list of URLs. We use its top-5-ranked URLs for the measurement, as users are more likely to click the top-5 results and AI-based answers usually only include around five references.
In total, we evaluated \textbf{3996} sources on \textbf{12} LLMs and search systems.

\begin{table*}[t!]\centering
\small
\caption{\textbf{Overall Evaluation Results}. Weighted gives the first three columns a 1.5 weight and scales the max to 100. Avg shows the average score of the eight metrics. Evaluating systems are first grouped by type and then ordered by their weighted overall score.}
\label{tab:overall-res}
\begin{tabular}{lrrrrrrrr|rr|c}\toprule
\textbf{System} & \textbf{\semantic} & \textbf{\fact} & \textbf{\obj} & \textbf{\fresh} & \textbf{\org} & \textbf{\auth} & \textbf{\domain} & \textbf{\ad} & \textbf{Avg}& \textbf{Weighted} & \textbf{Rank} \\\midrule
\textbf{GPT-5} & 3.923 & \textbf{4.772} & \textbf{4.547} & 4.490 & \textbf{4.793} & \textbf{4.433} & \textbf{4.736} & \textbf{4.016} & \textbf{4.462} & \textbf{89.081} & 1 \\
\textbf{Grok-4.1-Fast}& 4.328 & 4.508 & 3.961 & 4.180 & 4.315 & 3.993 & 4.334 & 3.633 & 4.153 & 83.381 & 2 \\
\textbf{GPT-4o} &4.241 &4.207 &3.925 & 4.524 &3.963 &4.150 &4.180 &3.344 & 4.067 & 81.518 & 4 \\
\textbf{Claude-Sonnet-4.5} & 4.209 & 4.217 & 3.818 & 4.447 & 4.178 & 4.202 & 4.079 & 3.336 & 4.061 & 81.282 & 5 \\
\textbf{Gemini-3-Pro} & 3.633 & 4.133 & 3.851 & \textbf{4.743} & 4.141 & 3.891 & 3.993 & 3.502 & 3.988 & 79.379 & 8 \\
\textbf{Gemini-3-Flash} & 3.633 & 4.100 & 3.925 & 4.459 & 4.097 & 4.057 & 3.962 & 3.416 & 3.960 & 78.981 & 9 \\
\textbf{Perplexity-Sonar-Pro} &3.691 & 4.075 &3.619 &4.358 &3.806 &4.220 &4.183 &3.643 &3.952 &78.547 & 10 \\
\textbf{Gemini-2.5-Flash} &3.520 &3.995 & 3.898 &4.445 &4.054 &4.163 &4.061 &3.336 &3.938 & 78.339 & 11 \\
\hline
\textbf{Gensee} & \textbf{4.435} & 4.432 & 3.895 & 4.343 & 3.970 & 3.970 & 4.102 & 3.343 & 4.060 & 81.795 & 3 \\
\textbf{Exa} &3.649 &4.120 & 4.024 &4.346 &4.125 & 4.306 &4.240 &3.391 &4.021 & 80.108 & 6 \\
\textbf{Tavily} &3.545 &4.066 & 3.801 &4.448 &4.043 &4.033 &4.156 &3.332 &3.933 & 78.268 & 12 \\
\hline
\textbf{Google-Search} &3.948 &4.222 &3.776 &4.059 &3.879 &4.133 & 4.232 & 3.749 & 4.000 & 79.939 & 7 \\
\bottomrule
\end{tabular}
\end{table*}
\begin{table*}[t!]\centering
\small
\caption{\textbf{Evaluation Results by Request Category.} Results shown are weighted score across metrics for each system and query type.}
\label{tab:overall-category}
\begin{tabular}{lrrrrr|r}
        \toprule
        \textbf{System} & \textbf{Argument} & \textbf{Reasoning} & \textbf{Factual} & \textbf{Social} & \textbf{Shopping} & \textbf{Overall} \\
        \midrule
        \textbf{GPT-5} & \textbf{90.192} & \textbf{88.337} & \textbf{88.995} & \textbf{87.840} & \textbf{85.553} & \textbf{89.116} \\
        \textbf{Grok-4.1-Fast} & 90.351 & 78.012 & 86.002 & 78.820 & 84.486 & 83.381 \\
        \textbf{GPT-4o} & 82.605 & 78.283 & 86.605 & 79.211 & 80.884 & 81.517 \\
         \textbf{Claude-Sonnet-4.5} & 79.398 & 79.158 & 85.560 & 74.398 & 80.104 & 81.281 \\
        \textbf{Gemini-3-Pro} & 81.421 & 72.632 & 81.873 & 72.372 & 81.355 & 79.359 \\
        \textbf{Gemini-3-Flash} & 81.053 & 72.285 & 80.402 & 74.257 & 78.730 & 78.905 \\
         \textbf{Perplexity-Sonar-Pro} & 82.516 & 71.139 & 80.994 & 75.700 & 79.391 & 78.501 \\
        \textbf{Gemini-2.5-Flash} & 82.126 & 73.860 & 77.674 & 74.407 & 81.018 & 78.269 \\
        \hline
        \textbf{Gensee} & 85.370 & 76.678 & 85.947 & 77.945 & 79.438 & 81.876 \\
        \textbf{Exa} & 86.446 & 70.831 & 83.593 & 76.966 & 81.975 & 80.204 \\
        \textbf{Tavily} & 83.560 & 68.339 & 82.274 & 73.671 & 78.808 & 78.170 \\
        \hline
        \textbf{Google-Search} & 85.275 & 75.338 & 83.531 & 76.224 & 77.055 & 79.939 \\
        \bottomrule
    \end{tabular}
\end{table*} 

\subsection{Results}

Table~\ref{tab:overall-res} summarizes our evaluation results, presenting the average scores across all 100 requests for each system under the eight metrics. We also include an average score and a weighted average score. The weighted average gives the content-based rubrics a higher weight than meta-attribute ones, with a ratio of 3:2. 
Table~\ref{tab:overall-category} details the performance across different query intents. 

Overall, \textbf{GPT-5} achieves the highest weighted overall score and ranks first in six out of eight metrics. GPT-5 also consistently ranks first across all five query types. GPT-5 excels at vetting sources for Organization, Author, and Domain Authority, indicating that it likely involves an internal screening process that choose more authoritative sources than other ones even when they have similar content. It also achieves the highest Factuality, NEutrality, and Layout Clarity scores, suggesting that more authoritative pages are likely to also be more vetted and objective (detailed analysis to be presented in Section~\ref{sec:metric-corr}).

\textbf{Grok-4.1-Fast} achieves the second-highest weighted overall score.
The performance of Grok-4.1-Fast is the most surprising, as we use its non-reasoning mode. Its Content Relevance score is much higher than GPT-5, and it performs reasonably well in all other metrics. We suspect that it has been fine-tuned to construct coherent, logical arguments even without reasoning chains. Looking at query types, Grok performs better in Shopping, Argumentative, and Factual types but worst in Reasoning, suggesting its strong capability of constructing arguments and real-time information but less robust on harder questions that require deeper thinking. 

\textbf{Gensee} search achieves the third-highest weighted overall score, mainly because of its overwhelmingly high Content-Relevance score. Different from Exa and Tavily search, Gensee has a ``Deep Search'' mode that involves internal reasoning steps that are tailored to web search, explaining its high Content-Relevance score. However, when selecting more relevant sources, it overlooks source authority, explaining its lower Organization, Author, and Domain Authority scores.

The lowest weighted overall scores are Perplexity-Sonar-Pro, Gemini-2.5-Flash, and Tavily search. 
\textbf{Perplexity-Sonar-Pro} suffers from a significant penalty in Neutrality (3.62), the lowest score among all. While it performs adequately on Freshness and Layout, its tendency to retrieve or synthesize content with lower neutrality suggests it may be over-indexing on opinionated sources or failing to balance conflicting viewpoints as effectively as GPT-5 (4.55). Additionally, it struggles with Informational queries (3.59), indicating that while it functions well for quick answers (Argument 4.24), it lacks the depth required for comprehensive topic exploration.

\textbf{Gemini-2.5-Flash} is the smallest LLM among all the LLMs. Its primary point of failure is Content Relevance (lowest of all systems), indicating its failure to identify sources that can answer a query properly. This could be due to its lack of a strong semantic reasoning capability to filter search results effectively.

\textbf{Tavily} ranks lowest overall among all systems evaluated. Despite it being an AI-native search tool, it failed to retrieve enough content-relevant, factual, or objective sources (among the lowest in these scores). Its performance on the Reasoning request type is also among the lowest. This indicates that it lacks strong reasoning capability. 

Interestingly, \textbf{Google}, being a pure SERP-style baseline, achieves the 7th among all systems. Counterintuitively, its Content Relevance score is higher than the overall-top-performing GPT-5. This indicates that even with just a single-step SERP-style retrieval, the sources retrieved can be of high relevance. It also achieves relatively high scores in Layout Clarity and Author Accountability. Surprisingly, its Freshness is the lowest among all systems, likely due to its less frequent re-indexing.

\boldunderparagraph{Insight 1:}
{\it{    The next leap of AI-based search should go for architectures that explicitly weight source credibility and content quality, mimicking the editorial judgment of a human expert rather than keyword matching machines.
}}

{
\begin{figure}
\begin{center}
\centerline{\includegraphics[width=\columnwidth]{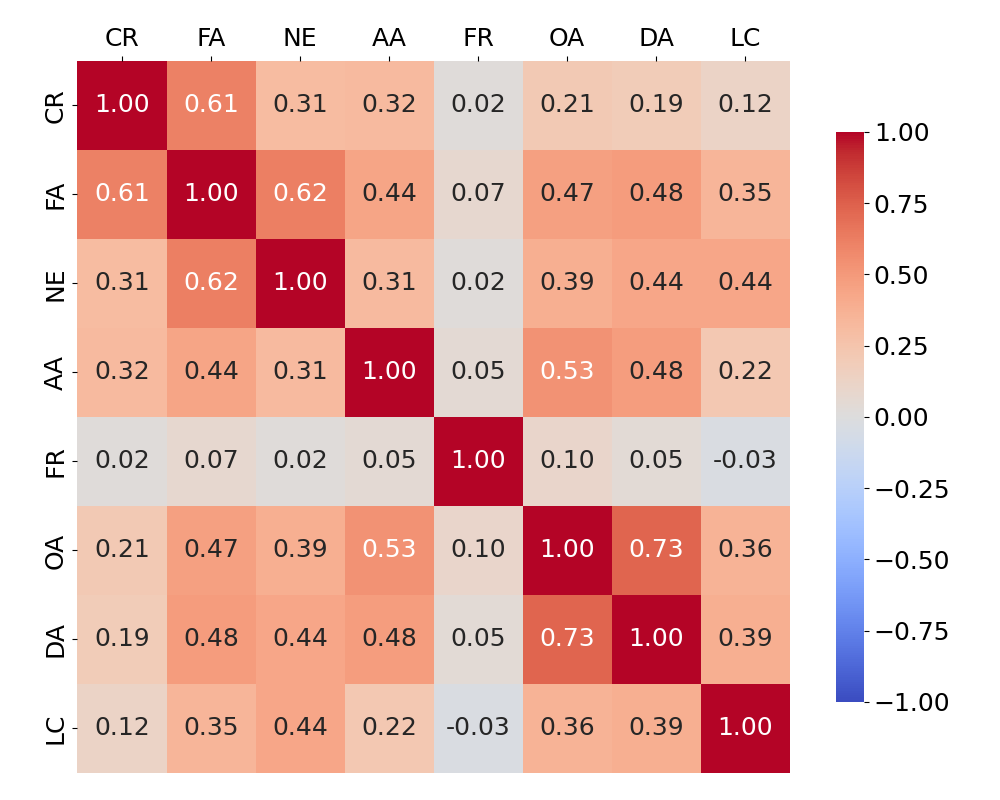}}
\caption{\textbf{Correlation between Metrics}. Heatmap of correlation between different metrics; results from all evaluating systems are aggregated.}
\label{fig-metric-correlation}
\end{center}
\end{figure}
}

\subsection{Correlation of Metrics}
\label{sec:metric-corr}
To understand these results and our metrics better, we performed a correlation analysis across the eight quality metrics. Figure~\ref{fig-metric-correlation} plots the heatmap of the Pearson correlations with aggregated data from all evaluating systems. Overall, there are two clusters of correlated metrics. First, the three types of source accountability metrics (\org, \auth, \domain) are correlated, with the strongest correlation between Ownership Accountability (\org) and Domain Authority (\domain). This  suggests that credible domains often clearly disclose their owners who are authoritative. 
The second cluster is among the three content-based scores (\semantic, \fact, \obj). As expected, the three content-related scores are correlated. More interestingly, Freshness (\fresh) exhibits near-zero correlation with all other metrics for all the systems, demonstrating that all systems evaluate the temporal aspect of content as an entirely independent variable, and that it has no relationship with either the content or other meta-features of a page.

\subsection{Understanding the Impact of Request Types}
\label{sec:req-type}

Performance variability across query types highlights different emphasis of query types. Commercial intent queries (VACOS) achieved the highest scores for Freshness and Factual Accuracy but the lowest for Layout Clarity. This reflects the nature of e-commerce environments, where high-utility, rapidly updated specification data is frequently embedded within visually cluttered, ad-heavy interfaces. Community-based inquiries (Quora) prioritized Freshness at the expense of Objectivity. In contrast, factuality-centric tasks (Pinocchios) inverted this pattern, maximizing the ``Trust'' metrics (organization, author, domain authority, and factuality) while having the lowest Freshness scores, indicating authoritative content may appear in older, established sources. Finally, multi-hop reasoning tasks (HotpotQA) exposed a critical limitation in all the systems, yielding the lowest Content Relevance scores.

\begin{table*}[!htp]\centering
\small
\caption{\textbf{HotpotQA (Reasoning) Results.} The answer score shown is exact match percentage to ground truth. The last column shows the weighted \bench\ overall scores. Rows ordered by answer score.}
\label{tab:hotpotqa}
\begin{tabular}{lr|rrrrrrrr|r}
\toprule
\textbf{System} & \textbf{Answer Score} & \textbf{\semantic} & \textbf{\fact} & \textbf{\obj} & \textbf{\fresh} & \textbf{\org} & \textbf{\auth} & \textbf{\domain} & \textbf{\ad} & \textbf{Weighted Source} \\
\midrule
\textbf{GPT-5} & \textbf{18.33\%} & 3.923 & \textbf{4.772} & \textbf{4.547} & 4.490 & \textbf{4.793} & \textbf{4.433} & \textbf{4.736} & \textbf{4.016} & \textbf{89.081} \\
\textbf{Gensee} & 14.04\% & \textbf{4.435} & 4.432 & 3.895 & 4.343 & 3.970 & 3.970 & 4.102 & 3.343 & 81.795 \\
\textbf{Grok-4.1-Fast} & 10.00\% & 4.328 & 4.508 & 3.961 & 4.180 & 4.315 & 3.993 & 4.334 & 3.633 & 83.381 \\
\textbf{Perplexity-Sonar-Pro} & 10.00\% & 3.691 & 4.075 & 3.619 & 4.358 & 3.806 & 4.220 & 4.183 & 3.643 & 78.547 \\
\textbf{Claude-Sonnet-4.5} & 9.26\% & 4.209 & 4.217 & 3.818 & 4.447 & 4.178 & 4.202 & 4.079 & 3.336 & 81.282 \\
\textbf{Gemini-3-Pro} & 8.33\% & 3.633 & 4.133 & 3.851 & \textbf{4.743} & 4.141 & 3.891 & 3.993 & 3.502 & 79.379 \\
\textbf{Gemini-3-Flash} & 8.33\% & 3.633 & 4.100 & 3.925 & 4.459 & 4.097 & 4.057 & 3.962 & 3.416 & 78.981 \\
\textbf{Gemini-2.5-Flash} & 6.67\% & 3.520 & 3.995 & 3.898 & 4.445 & 4.054 & 4.163 & 4.061 & 3.336 & 78.339 \\
\textbf{GPT-4o} & 5.00\% & 4.241 & 4.207 & 3.925 & 4.524 & 3.963 & 4.150 & 4.180 & 3.344 & 81.518 \\
\bottomrule
\end{tabular}
\end{table*}
{
\begin{figure}
\begin{center}
\centerline{\includegraphics[width=\columnwidth]{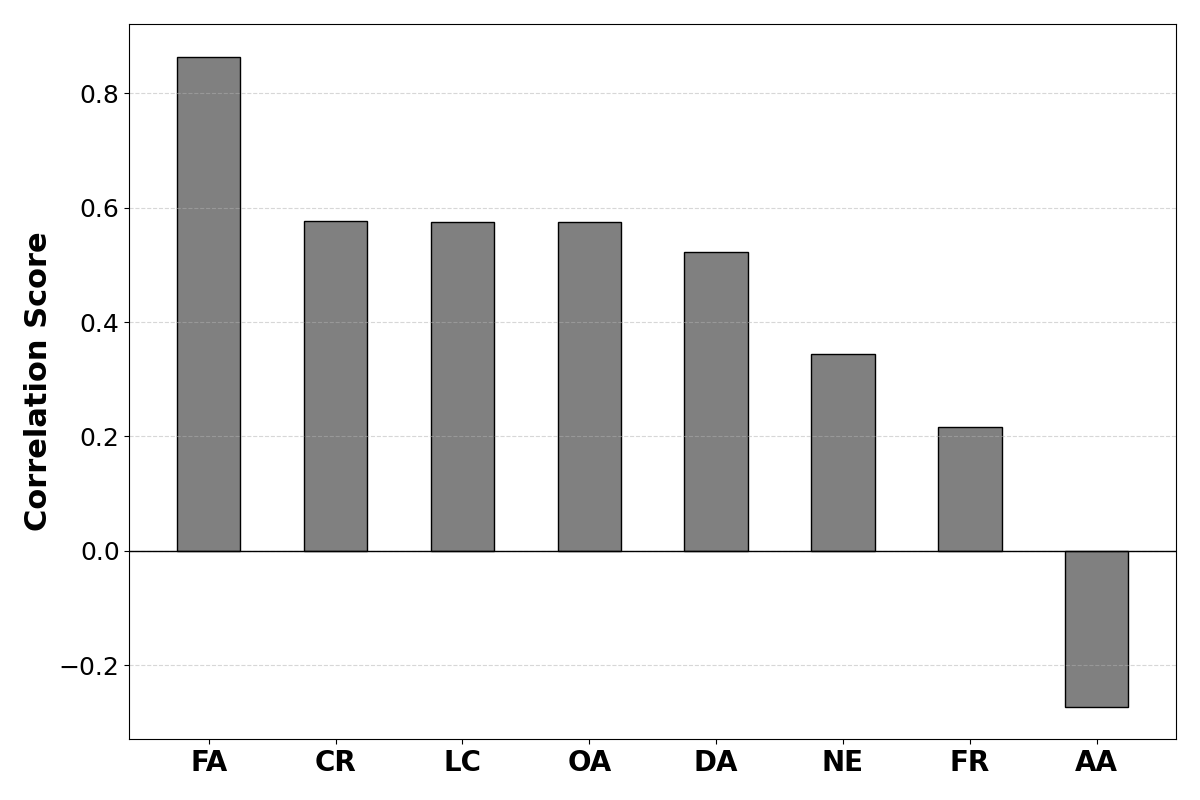}}
\caption{\textbf{Correlation in HotpotQA}.}
\label{fig-hotpotqa}
\end{center}
\end{figure}
}

\section{Deep Dive into GenAI and Web Search}

To further understand the relationship between web source quality and AI answer quality and to discern the impact of LLMs vs. search tools, we conduct a set of deep-dive experiments to strengthen our findings.

\subsection{Source Quality and Answer Quality}
To understand web source quality (\ie, \bench's scores) relate to the text answer quality (\ie, original benchmark used in Table~\ref{tbl:query}), we use HotpotQA (reasoning) and DebateQA (argumentative)'s original evaluators to collect the text-answer scores for the AI systems we evaluate. As Google, Exa, and Tavily do not return a text answer, we do not include them in this evaluation.

Table~\ref{tab:hotpotqa} shows the answer score rated using HotpotQA's ground truth answer by exact match percentage, together with the eight metrics in \bench\ and the weighted overall \bench\ score. Figure~\ref{fig-hotpotqa} plots the correlation between each metric and the answer score. As expected, Factuality has a high positive correlation with the answer score. While most other metrics also have positive correlation, Author Accountability has a negative correlation with the answer score. In practice, users are more inclined to be willing to check sources from well-established authors. This implies that traditional answer-only evaluation is insufficient.

\begin{table*}[t!]\centering
\small
\caption{\textbf{DebateQA Results.} Rows are ordered by the weighted score ranking from the provided dataset. PD (Perspective Diversity) and DA (Dispute Awareness) are prepended from the previous analysis.}
\label{tab:debateqa}
\begin{tabular}{lrr|rrrrrrrr|r}
\toprule
\textbf{System} & \textbf{PD} $\downarrow$ & \textbf{DA} $\uparrow$ & \textbf{\semantic} & \textbf{\fact} & \textbf{\obj} & \textbf{\fresh} & \textbf{\org} & \textbf{\auth} & \textbf{\domain} & \textbf{\ad} & \textbf{Weighted Source} \\
\midrule
\textbf{GPT-5} & \textbf{3.621} & 0.383 & 3.923 & \textbf{4.772} & \textbf{4.547} & 4.490 & \textbf{4.793} & \textbf{4.433} & \textbf{4.736} & \textbf{4.016} & \textbf{89.081} \\
\textbf{Claude-Sonnet-4.5} & 3.800 & 0.633 & 4.209 & 4.217 & 3.818 & 4.447 & 4.178 & 4.202 & 4.079 & 3.336 & 81.282 \\
\textbf{Gemini-3-Flash} & 3.885 & \textbf{0.717} & 3.633 & 4.100 & 3.925 & 4.459 & 4.097 & 4.057 & 3.962 & 3.416 & 78.981 \\
\textbf{Gensee} & 3.893 & 0.233 & \textbf{4.435} & 4.432 & 3.895 & 4.343 & 3.970 & 3.970 & 4.102 & 3.343 & 81.795 \\
\textbf{Perplexity-Sonar-Pro} & 3.901 & 0.483 & 3.691 & 4.075 & 3.619 & 4.358 & 3.806 & 4.220 & 4.183 & 3.643 & 78.547 \\
\textbf{GPT-4o} & 3.907 & 0.526 & 4.241 & 4.207 & 3.925 & 4.524 & 3.963 & 4.150 & 4.180 & 3.344 & 81.518 \\
\textbf{Gemini-2.5-Flash} & 3.961 & 0.450 & 3.520 & 3.995 & 3.898 & 4.445 & 4.054 & 4.163 & 4.061 & 3.336 & 78.339 \\
\textbf{Grok-4.1-Fast} & 4.069 & 0.317 & 4.328 & 4.508 & 3.961 & 4.180 & 4.315 & 3.993 & 4.334 & 3.633 & 83.381 \\
\textbf{Gemini-3-Pro} & 4.088 & 0.583 & 3.633 & 4.133 & 3.851 & \textbf{4.743} & 4.141 & 3.891 & 3.993 & 3.502 & 79.379 \\
\bottomrule
\end{tabular}
\end{table*}

We also evaluate the text answers generated for DebateQA (Argumentative) using DebateQA's evaluator. We use DebateQA's evaluator to get two metrics they use: Perspective Diversity (PD, lower the better) and Dispute Awareness (DA, higher the better). These results and \bench's scores are summarized in Table~\ref{tab:debateqa}. As seen, the overall source quality largely determines the Perspective Diversity metric.
However, high content diversity does not inherently guarantee Dispute Awareness.

\boldunderparagraph{Insight 2:}
{\it{Web source quality highly affects AI answer quality, but today's AI benchmarks fail to evaluate all aspects of source quality.
}}

\begin{table*}[h!]
\centering
\small
\renewcommand{\arraystretch}{0.75} 
\setlength{\tabcolsep}{4pt} 

\caption{\textbf{GenAI and Search Result Overlapping}}
\label{tab:se-ge}
\begin{tabular}{lccccccccccc}
\toprule
\textbf{Metric} & \textbf{GPT-5} & \textbf{GPT-4o} & \textbf{Grok-4.1} & \textbf{Gem-3-pr} & \textbf{Gem-3-Fl} & \textbf{Gem-2.5-Fl} & \textbf{Plxty} & \textbf{Exa} & \textbf{Tavily} & \textbf{Gensee} & \textbf{Claude} \\
\midrule
\% in Google & 15.99 & 27.53 & 29.67 & 20.95 & 23.5 & 31.96 & 40 & 44.66 & 55.45 & 28.4 & 37.1 \\
W. Source & 89.081 & 81.518 & 83.381 & 79.379 & 78.981 & 78.339 & 78.547 & 80.108 & 78.268 & 81.795 & 81.282 \\
\bottomrule
\end{tabular}
\end{table*}

\subsection{Search Engine and Generative Engine}

An interesting question is to understand the relationship between GenAI (semantic search) and SERP (keyword search) and how that relates to the overall source quality. 
To understand this, we analyze the number of sources that each AI-based solution returns that are also in the first five pages of Google search results. Table~\ref{tab:se-ge} shows this result. Interestingly, GPT-5, the best-performing system in \bench, has the lowest amount of sources appearing in Google search results, and Tavily, the worst-performing in \bench, has the highest overlap with Google search. This indicates that good sources are mostly not in the one-shot, keyword-based searches, justifying the importance of AI solutions in the new era of information search.

\boldunderparagraph{Insight 3:}
{\it{    AI search is not merely summarizing SERP results, but actively discovering high-quality buried evidence that require deeper thinking.
}}

\begin{table*}[!htp]\centering
\small
\caption{\textbf{DeepSeek Results.}}
\label{tab:deepseek}
\begin{tabular}{lrrrrrrrr|c}
\toprule
\textbf{System} & \textbf{\semantic} & \textbf{\fact} & \textbf{\obj} & \textbf{\fresh} & \textbf{\org} & \textbf{\auth} & \textbf{\domain} & \textbf{\ad} & \textbf{Weighted Avg} \\ \midrule
\textbf{DS-Chat-Tavily} & 3.783 & 4.087 & 3.783 & 4.217 & 4.232 & 3.855 & 4.246 & 3.507 & 70.085 \\
\textbf{DS-Chat-Gensee} & \textbf{4.244} & 4.564 & \textbf{4.154} & \textbf{4.423} & 4.321 & 3.987 & 4.462 & \textbf{3.922} & \textbf{75.989} \\
\textbf{DS-Reasoning-Tavily} & 4.190 & \textbf{4.603} & 4.034 & 4.397 & \textbf{4.552} & \textbf{4.103} & \textbf{4.655} & 3.724 & 75.823 \\
\textbf{DS-Reasoning-Gensee} & 4.040 & 4.467 & 3.987 & 4.351 & 4.230 & 4.081 & 4.392 & 3.797 & 74.103 \\ \bottomrule
\end{tabular}
\end{table*}

\subsection{LLM and Search Tool}

To understand the importance of LLM vs. web search in delivering high-quality web references, we perform a controlled experiment with DeepSeek, which does not by itself have any web search capability but allows the supply of any external search tools. We feed the highest and lowest search tools found in our experiments, Gensee and Tavily, to two DeepSeek models, DeepSeek-3.2-Chat (no reasoning capability) and DeepSeek-3.2-Reason~\cite{deepseek_v3_2_2025}.
Table~\ref{tab:deepseek} shows the overall results.

Surprisingly, DeepSeek-Chat plus Gensee performs the best, with DeepSeek-Reasoning plus Tavily being the second best and DeepSeek-Reasoning plus Gensee being the third. 
These data suggest that high-quality web search acts as a functional substitute for advanced reasoning and even outperforms it. Interestingly, the DeepSeek-Reason model demonstrates its value better when paired with the lower-quality Tavily search than with the higher-quality Gensee search. This is likely due to Gensee search's internal thinking capability conflicts with some of DeepSeek's reasoning.

We further analyze all the internal steps of a sampled shopping query that looks for a laptop that satisfy multiple requirements. Tavily returns more biased sources and sources that focus on one requirement, and DeepSeek-Chat takes the returned sources without further thinking. Gensee returns more well-rounded sources covering all requirements, leading to a higher-quality final answer and references. The Tavily issue, while not fixed by Deepseek-Chat, is captured and fixed by Deepseek-Reason, as it finds the misalignment of search results with the user answer and explictly asks Tavily to search for previously unincluded requirements.

\boldunderparagraph{Insight 4:}
{\it{Instead of relying on a model to ``think'' its way through noise`, providing superior, well-curated context allows simpler models to achieve better outcomes.}}

\section{Conclusion}

\section*{Impact Statement}

This paper presents work whose goal is to advance the field of Machine
Learning. There are many potential societal consequences of our work, none which we feel must be specifically highlighted here.

\bibliography{local}
\bibliographystyle{icml2026}

\newpage
\appendix
\onecolumn




\appendix
\onecolumn
\section{Generative Engine Source Reliability Rubric}
\label{appendix:rubric}

In this section, we present the revised evaluation rubric used to assess the reliability of sources retrieved by Generative Engines. This rubric synthesizes findings from recent literature \citep{true_framework, decoding_ai_judgment} and industry standards \citep{google_quality_guidelines} with specific implementation guidance for automated scoring.


{
\small 
\renewcommand{\arraystretch}{1.4} 
\begin{longtable}{p{0.20\textwidth} p{0.38\textwidth} p{0.37\textwidth}}

\caption{Revised Source Reliability Rubric with Citations \& Guidance. Dimensions are scored on a 1--5 scale.} \label{tab:rubric_revised} \\
\toprule
\textbf{Criterion} & \textbf{Rationale, Citations \& Guidance} & \textbf{Scoring Guidelines (1--5 Scale)} \\
\midrule
\endfirsthead

\multicolumn{3}{c}{\textit{... continued from previous page}} \\
\toprule
\textbf{Criterion} & \textbf{Rationale, Citations \& Guidance} & \textbf{Scoring Guidelines (1--5 Scale)} \\
\midrule
\endhead

\midrule
\multicolumn{3}{r}{\textit{Continued on next page ...}} \\
\endfoot

\bottomrule
\endlastfoot

\multicolumn{3}{c}{\cellcolor{groupgray}\textbf{Group 1: Content Assessment}} \\
\midrule

\textbf{1. Semantic Relevance} & 
\textit{Core Utility.} \citet{true_framework} validated that ``Intent Alignment'' is distinct from factual accuracy. \citet{pinterest_ranking} showed fine-grained scales outperform binary metrics. \newline \newline
\textbf{Implementation Guidance:} \newline Use \texttt{Text\_content\_length} as a signal. If length is $<$300 chars, penalize relevance unless it is a perfectly concise factual answer. & 
\textbf{5 (Fully Meets):} Comprehensive, direct answer. No further searching needed. \newline
\textbf{4 (Highly Meets):} Highly relevant but lacks minor details or slightly outdated. \newline
\textbf{3 (Moderately Meets):} Broadly on topic but misses specific intent; requires more clicks. \newline
\textbf{2 (Slightly Meets):} Tangential keyword match only. Misses core question. \newline
\textbf{1 (Fails to Meet):} Irrelevant, wrong language, or technical failure. \\ 
\addlinespace[10pt]

\textbf{2. Factual Accuracy} & 
\textit{Evidence Verification.} \citet{Li2024} identify lack of citations as a primary cause of hallucination. \citet{decoding_ai_judgment} found accuracy is the strongest predictor of credibility. \newline \newline
\textbf{Note:} Primary sources (Gov, Edu, Science) are weighted higher than secondary sources (News). & 
\textbf{5 (Verified):} Claims cited with primary sources (Gov, Edu, Science). \newline
\textbf{4 (Accurate):} Aligns with consensus. Secondary citations (News) exist. \newline
\textbf{3 (Unverified):} Plausible ``common knowledge'' but lacks evidence/links. \newline
\textbf{2 (Suspect):} Fringe theories, logic gaps, or context omissions. \newline
\textbf{1 (False):} Demonstrably false, dangerous, or debunked. \\ 
\addlinespace[10pt]

\textbf{3. Objectivity \& Tone} & 
\textit{Bias Detection.} \citet{decoding_ai_judgment} identify ``Language Manipulation'' and emotional charging as key signals for low-credibility domains. & 
\textbf{5 (Neutral):} Clinical/Academic. Multiple viewpoints. Zero emotional loading. \newline
\textbf{4 (Balanced):} Professional journalistic tone. Acknowledges counter-arguments. \newline
\textbf{3 (Opinionated):} Clear subjective bias/op-ed but grounded in facts. \newline
\textbf{2 (Inflammatory):} Emotional triggers (``Shocking!''), excessive punctuation. \newline
\textbf{1 (Propaganda):} Hate speech, fear-mongering, incoherent. \\ 
\midrule

\multicolumn{3}{c}{\cellcolor{groupgray}\textbf{Group 2: Source \& Metadata Assessment}} \\
\midrule

\textbf{4. Information Freshness} & 
\textit{Utility Blocker.} \citet{hoh_benchmark} prove ``Timeliness Awareness'' is critical for RAG utility. \newline \newline
\textbf{Waterfall Logic Implementation:} \newline
\textbf{1. Priority 1:} Check \texttt{last\_updated}. \newline
\hspace*{1em} If $\le$ 1 year from Jan 2026 $\to$ Score 5. \newline
\textbf{2. Priority 2:} If P1 is null, use \texttt{date}. \newline
\textbf{3. Priority 3:} If both null, use text markers (e.g., ``RTX 50-series''). \newline
\textit{Exception:} If \texttt{last\_updated} (2025) $\gg$ \texttt{date} (2012), treat as Score 5. & 
\textbf{5 (Current):} Dated within last 6--12 months. Data is current. \newline
\textbf{4 (Recent):} 1--2 years old but valid. \newline
\textbf{3 (Static):} Undated ``evergreen''. References older context (e.g., ``Windows 10''). \newline
\textbf{2 (Outdated):} Clearly obsolete (old laws, discontinued products). \newline
\textbf{1 (Legacy):} Ancient ($>$10 years) and factually wrong due to time. \\ 
\addlinespace[10pt]

\textbf{5. Author Accountability} & 
\textit{Entity Verification.} \citet{decoding_ai_judgment} highlight bios as top credibility features. Explicitly required by \citet{google_quality_guidelines}. & 
\textbf{5 (Expert):} Named author with verifiable bio, credentials linked (MD, PhD). \newline
\textbf{4 (Professional):} Named author (Staff Writer) or reputable group (Editorial Board). \newline
\textbf{3 (Opaque):} ``Admin'' or ``Team'' authorship. \newline
\textbf{2 (Hidden):} No author listed where expected. \newline
\textbf{1 (Deceptive):} Fake persona or AI-generated face. \\ 
\addlinespace[10pt]

\textbf{6. Ownership Transparency} & 
\textit{Legitimacy.} \citet{decoding_ai_judgment} found declaring funding/owners is a significant differentiator for high-quality sites. \newline \newline
\textbf{The ``Null = Penalty'' Rule:} Missing ownership info results in automatic downgrading. & 
\textbf{5 (Fully Transparent):} Physical address, phone number, and clear leadership/funding disclosed. \newline
\textbf{4 (Partially Transparent):} Clear ``About'' desc. and leadership names, but missing physical contact info. \newline
\textbf{3 (Opaque):} Contact form only; vague ``About'' section; no specific names/funding. \newline
\textbf{2 (Anonymous Org):} No ``About'' info or ownership details. \newline
\textbf{1 (Hidden):} Intentional shell company signs or active attempts to hide ownership. \\ 
\addlinespace[10pt]

\textbf{7. Domain Authority} & 
\textit{Reputation Proxy.} \citet{google_quality_guidelines} emphasize ``Reputation Research'' (reviews, awards) as a core trust signal. \newline \newline
\textbf{Implementation Guidance:} \newline
If \texttt{Domain\_category} is \texttt{.gov} or \texttt{.edu}, automatically assign Score 5. & 
\textbf{5 (Authority):} Gov/Edu, Major News, established brand ($>$10 years). \newline
\textbf{4 (Trusted):} Established commercial brand or niche expert. \newline
\textbf{3 (Unknown):} Functional blog/small biz. No reputation history. \newline
\textbf{2 (Low Quality):} Content farm, parasite SEO, clickbait network. \newline
\textbf{1 (Blacklisted):} Known scam, phishing, or disinformation. \\ 
\midrule

\multicolumn{3}{c}{\cellcolor{groupgray}\textbf{Group 3: Technical / UX Assessment}} \\
\midrule

\textbf{8. Layout \& Ad Density} & 
\textit{User Experience.} \citet{geo_bench} demonstrate that DOM tree complexity and ad density negatively correlate with quality. \newline \newline
\textbf{Implementation Guidance:} \newline Look for ``ad-interstitial'' labels or fragmented structures. High volume of affiliate links suggests Score 2--3. & 
\textbf{5 (Clean):} Zero intrusive ads. High skimmability. \newline
\textbf{4 (Standard):} Banners/sidebars that do not block content. \newline
\textbf{3 (Cluttered):} Auto-playing video, sticky footers, heavy monetization. \newline
\textbf{2 (Obstructive):} Pop-ups require dismissal. ``Chumbox'' fake articles at bottom. \newline
\textbf{1 (Spam/Unusable):} Overlays, broken layout, malicious redirects. \\ 

\end{longtable}
}


\section{LLM Evaluator Prompts}
\label{appendix:prompts}

This appendix details the exact prompts used to automate the grading process. The evaluation was split into three distinct calls to minimize context window interference and group related reasoning tasks. 

\subsection{Group 1: Content Quality Assessment Prompts}
\label{appendix:prompt_content}
\begin{longtable}{|p{0.2\linewidth}|p{0.75\linewidth}|}
\hline
\textbf{Prompt Type} & \textbf{Content} \\
\hline

\textbf{System Prompt} & 
You are a research evaluator specializing in Information Retrieval (IR). Your task is to perform an internal quality assessment of a cited website's text content relative to a specific user query. \\
\hline

\textbf{User Prompt Template} & 
\textbf{Rubrics:} \newline
\textbf{1. Semantic Relevance (1-5)} \newline
5 (Fully Meets): Comprehensive, direct answer. No further searching needed. \newline
4 (Highly Meets): Highly relevant but lacks minor details or slightly outdated. \newline
3 (Moderately Meets): Broadly on topic but misses specific intent; requires more clicks. \newline
2 (Slightly Meets): Tangential keyword match only. Misses core question. \newline
1 (Fails to Meet): Irrelevant, wrong language, or technical failure. \newline
\textit{Implementation Guidance: Use text\_content\_length as a signal; if length is <300 characters, penalize relevance unless it is a perfectly concise factual answer.}

\vspace{0.2cm}
\textbf{2. Factual Accuracy \& Citations (1-5)} \newline
5 (Verified): Claims cited with primary sources (Gov, Edu, Science). \newline
4 (Accurate): Aligns with consensus. Secondary citations (News) exist. \newline
3 (Unverified): Plausible ``common knowledge'' but lacks evidence/links. \newline
2 (Suspect): Fringe theories, logic gaps, or context omissions. \newline
1 (False): Demonstrably false, dangerous, or debunked.

\vspace{0.2cm}
\textbf{3. Objectivity \& Tone (1-5)} \newline
5 (Neutral): Clinical/Academic. Multiple viewpoints. Zero emotional loading. \newline
4 (Balanced): Professional journalistic tone. Acknowledges counter-arguments. \newline
3 (Opinionated): Clear subjective bias/op-ed but grounded in facts. \newline
2 (Inflammatory): Emotional triggers (``Shocking!''), excessive punctuation. \newline
1 (Propaganda): Hate speech, fear-mongering, incoherent.

\rule{\linewidth}{0.5pt}

\textbf{Few-Shot Calibration Examples} \newline
\textbf{Example 1: The Expert Document (Query 81 - Apache Mahout)} \newline
\textit{Query:} How do I evaluate AUC in Apache Mahout? \newline
\textit{Analysis:} Relevance: 5 (Perfect intent/class alignment); Accuracy: 5 (Official docs); Tone: 3 (Clinical but saturated with boilerplate).

\vspace{0.1cm}
\textbf{Example 2: The GEO/Thin Content (Query 11 - Trenchcoats)} \newline
\textit{Query:} Reasonably-priced (under \$50) short trenchcoat recommendations. \newline
\textit{Analysis:} Relevance: 3 (Matches keyword but fails specific ``short'' constraint); Accuracy: 3 (Common knowledge, no links); Tone: 4 (Professional but slightly promotional).

\vspace{0.1cm}
\textbf{Example 3: The High-Quality Review (Query 1 - Laptops)} \newline
\textit{Query:} Smaller laptop for older games/word processing, SSD preferred. \newline
\textit{Analysis:} Relevance: 4 (Highly relevant, explicit SSD mention missing in summary); Accuracy: 5 (Lab testing); Tone: 4 (Journalistic but contains affiliate links).

\rule{\linewidth}{0.5pt}

\textbf{Task \& Output Format} \newline
Provide a score (1-5) and a short justification for each of the three metrics above. \newline
\textit{Input Data Packet:} \texttt{\{ "query\_id": "...", "query\_text": "...", "source\_url": "...", "source\_text": "...", "text\_length": "..." \}} \newline

\textit{Output Format (Return ONLY JSON):} \newline
\texttt{\{ "semantic\_relevance": \{ "score": int, "justification": "string" \}, "factual\_accuracy": \{ "score": int, "justification": "string" \}, "objectivity\_tone": \{ "score": int, "justification": "string" \} \}} \\
\hline

\end{longtable}

\newpage
\subsection{Group 2: Source Metadata Assessment Prompts}
\label{appendix:prompt_metadata}

\begin{longtable}{|p{0.2\linewidth}|p{0.75\linewidth}|}
\hline
\textbf{Prompt Type} & \textbf{Content} \\
\hline
\endhead 

\textbf{System Prompt} & 
You are a digital forensic auditor. Your task is to evaluate the external credibility and timeliness of a source based on its metadata and structural markers. \\
\hline


\textbf{User Prompt Template} & 
\textbf{Rubrics:}

\textbf{1. Information Freshness (1-5)} \newline
5 (Current): Dated within last 6-12 months. Data is current. \newline
4 (Recent): 1-2 years old but valid. \newline
3 (Static): Undated ``evergreen''. References older context. \newline
2 (Outdated): Clearly obsolete (old laws, discontinued products). \newline
1 (Legacy): Ancient ($>$10 years) and factually wrong. \newline
\textit{Implementation Guidance:} 1. Priority 1: \texttt{search\_result\_last\_updated} ($\le$ 1 yr from Jan 2026 = 5). 2. Priority 2: \texttt{search\_result\_date}. 3. Priority 3: Text Cues. \\

 & 
\textbf{2. Author Accountability (1-5)} \newline
5 (Expert): Named author with verifiable bio/credentials linked. \newline
4 (Professional): Named author (Staff Writer) or reputable group. \newline
3 (Opaque): ``Admin'' or ``Team'' authorship. \newline
2 (Hidden): No author listed where expected. \newline
1 (Deceptive): Fake persona or AI-generated face. \\

 & 
\textbf{3. Ownership Transparency} \newline
5 (Fully Transparent): Physical address, phone, clear leadership. \newline
4 (Partially Transparent): Clear ``About'', leadership names, no physical info. \newline
3 (Opaque): Contact form only; vague ``About''. \newline
2 (Anonymous Org): No ``About'' info or ownership details. \newline
1 (Hidden): Intentional shell company signs. \\

 & 
\textbf{4. Domain Authority (1-5)} \newline
5 (Authority): Gov/Edu, Major News, established brand. \newline
4 (Trusted): Established commercial brand or niche expert. \newline
3 (Unknown): Functional blog/small biz. \newline
2 (Low Quality): Content farm, parasite SEO. \newline
1 (Blacklisted): Known scam/disinformation. \\

 & 
\rule{\linewidth}{0.5pt} \newline
\textbf{Few-Shot Examples for Calibration} \newline
\textbf{Ex 1 (Apache Mahout):} Freshness: 5 (Dec 2025); Accountability: 4 (Apache Fdn); Transparency: 5; Domain: 5. \newline
\textbf{Ex 2 (Democracy In Action):} Freshness: 2 (2021); Accountability: 3; Transparency: 4; Domain: 4. \newline
\textbf{Ex 3 (PolitiFact):} Freshness: 3 (Active ecosystem); Accountability: 5; Transparency: 5; Domain: 4. \newline
\textbf{Ex 4 (Yoga Detox):} Freshness: 3 (Evergreen); Accountability: 5; Transparency: 5; Domain: 4. \\

 & 
\rule{\linewidth}{0.5pt} \newline
\textbf{Task \& Output Format} \newline
Provide a score (1-5) and justification for the four metrics. \newline
\textit{Input Data Packet:} \texttt{\{ "query\_id": "...", "current\_date": "Jan 25, 2026", ... \}}

\vspace{0.2cm}
\textit{Output Format (JSON Only):} \newline
\texttt{\{ "freshness": \{ "score": int, "justification": "string" \}, "author\_accountability": \{ "score": int, "justification": "string" \}, "ownership\_transparency": \{ "score": int, "justification": "string" \}, "domain\_authority": \{ "score": int, "justification": "string" \} \}} \\
\hline

\end{longtable}

\subsection{Group 3: Technical \& Ad Density Assessment Prompts}
\label{appendix:prompt_ads}

\begin{longtable}{|p{0.2\linewidth}|p{0.75\linewidth}|}
\hline
\textbf{Prompt Type} & \textbf{Content} \\
\hline
\endhead

\textbf{System Prompt} & 
You are a User Experience (UX) analyst. Your task is to evaluate the usability and monetization intensity of a website based on its extracted text structure. \\
\hline


\textbf{User Prompt Template} & 
\textbf{Rubrics:} \newline
\textbf{Layout \& Ad Density (1-5)} \newline
5 (Clean): Zero intrusive ads. High skimmability. \newline
4 (Standard): Banners/sidebars that do not block content. \newline
3 (Cluttered): Auto-playing video, sticky footers, heavy monetization. \newline
2 (Obstructive): Pop-ups require dismissal. ``Chumbox'' fake articles at bottom. \newline
1 (Spam/Unusable): Overlays, broken layout, malicious redirects. \newline
\textit{Implementation Guidance:} Since you are viewing extracted text, look for ``ad-interstitial text'' (e.g., ``ADVERTISEMENT'' labels) or fragmented text structures. High text volume containing frequent repetitive affiliate links (e.g., ``Check Price on Amazon'') suggests a Score 2 or 3. \\

 & 
\rule{\linewidth}{0.5pt} \newline
\textbf{Few-Shot Calibration Examples}

\vspace{0.1cm}
\textbf{Example 1: The Obstructive Media (Query 1 - Laptops)} \newline
\textit{Query:} Best ultraportable laptops for 2026? \newline
\textit{Source:} pcmag.com \newline
\textit{Analysis:} Layout \& Ad Density: 2 - Text is fragmented with frequent ``ADVERTISEMENT'' markers and ghost text from auto-play videos. Over 20 distinct commercial CTAs (``GET IT NOW'') exceed the threshold for high-intensity monetization.

\vspace{0.1cm}
\textbf{Example 2: The Cluttered Portal (Query 21 - Yoga Detox)} \newline
\textit{Query:} Can yoga really detoxify your body? \newline
\textit{Source:} healthgrades.com \newline
\textit{Analysis:} Layout \& Ad Density: 2 - Heavily saturated with repetitive navigational artifacts (``Find a Doctor'', ``Sign In'') and commercial CTAs. UX is secondary to user acquisition, triggering the ``Obstructive'' score.

\vspace{0.1cm}
\textbf{Example 3: The Clean Doc (Query 81 - Apache)} \newline
\textit{Query:} How do I evaluate AUC in Apache Mahout? \newline
\textit{Source:} mahout.apache.org \newline
\textit{Analysis:} Layout \& Ad Density: 5 - Continuous, logically structured technical document. Functional navigation (``SKIP NAVIGATION LINKS'') is present but standard; zero commercial markers or pop-ups. \\

 & 
\rule{\linewidth}{0.5pt} \newline
\textbf{Task \& Output Requirement} \newline
Provide a score (1-5) and a short justification for this metric.

\vspace{0.1cm}
\textit{Input Data Packet:} \newline
\texttt{\{ "source\_url": "...", "source\_text": "...", "text\_length": "..." \}}

\vspace{0.1cm}
\textit{Output Requirement (Return ONLY JSON):} \newline
\texttt{\{ "layout\_ad\_density": \{ "score": int, "justification": "string" \} \}} \\
\hline

\end{longtable}

\end{document}